\documentclass[10pt]{article} 
\usepackage[preprint]{tmlr}

\usepackage{amsmath,amsfonts,bm}

\def\eqref#1{equation~\ref{#1}}

\def\1{\bm{1}}

\DeclareMathAlphabet{\mathsfit}{\encodingdefault}{\sfdefault}{m}{sl}
\SetMathAlphabet{\mathsfit}{bold}{\encodingdefault}{\sfdefault}{bx}{n}

\DeclareMathOperator*{\argmin}{arg\,min}

\usepackage{hyperref}
\usepackage{url}

\usepackage{microtype} 
\usepackage{graphicx}
\usepackage{booktabs} 
\usepackage{multirow}       

\usepackage[utf8]{inputenc} 
\usepackage[T1]{fontenc}    
\usepackage{nicefrac}       
\usepackage{xcolor}         

\usepackage{mathtools}
\usepackage{amsthm}

\usepackage[capitalize,noabbrev]{cleveref}

\usepackage{subcaption}
\usepackage{pifont}
\newcommand{\cmark}{\ding{51}}%
\newcommand{\xmark}{\ding{55}}%
\usepackage{enumitem}

\title{Video Representation Learning with Joint-Embedding Predictive Architectures}

\author{\name Katrina Drozdov \email kve216@nyu.edu \\
      \addr Center for Data Science\\
      New York University
      \AND
      \name Ravid Shwartz-Ziv \email rs8020@nyu.edu  \\
      \addr  Center for Data Science\\
      New York University
      \AND
      \name Yann LeCun \email yann@cs.nyu.edu\\
      \addr Center for Data Science and Courant Institute\\
      New York University \\
      Meta FAIR}

\def\month{08}  
\def\year{2024} 
\def\openreview{\url{https://openreview.net/forum?id=XXXX}} 

\begin{document}

\maketitle

\begin{abstract}
Video representation learning is an increasingly important topic in machine learning research. We present \textbf{V}ideo \textbf{J}EPA  with \textbf{V}ariance-\textbf{C}ovariance \textbf{R}egularization (VJ-VCR): a joint-embedding predictive architecture for self-supervised video representation learning that employs variance and covariance regularization to avoid representation collapse. We show that hidden representations from our VJ-VCR contain abstract, high-level information about the input data. Specifically, they outperform representations obtained from a generative baseline on downstream tasks that require understanding of the underlying dynamics of moving objects in the videos. Additionally, we explore different ways to incorporate latent variables into the VJ-VCR framework that capture information about uncertainty in the future in non-deterministic settings.  

\end{abstract}

\section{Introduction}
The rapid increase in video data across various domains has created a pressing need for effective video representation learning methods that automatically extract and encode the essential elements of video content into compact and informative features. In particular, the goal behind video representation learning is to develop machine learning models that efficiently interpret complex, high-dimensional visual information by capturing key aspects unique to video data such as motion, scene context, and temporal dynamics. This capability is crucial for applications that require real-time understanding of dynamic environments such as robotic navigation, where robots must maneuver safely in unpredictable surroundings \citep{nahavandi2022comprehensive}; healthcare, where continuous video analysis can assist in medical diagnostics \citep{asan2014using}; and autonomous driving, which relies on accurate perception of the road and its surroundings to ensure safe operation \citep{chen2024end}. As these applications continue to evolve, robust video representations are essential for enabling reliable, responsive, and intelligent systems.

While powerful, traditional supervised learning approaches to video representation learning require vast amounts of labeled data, which is often expensive to obtain. Self-supervised learning (SSL) for video provides a promising alternative, where models learn to understand video content without relying on external annotations. SSL approaches typically involve designing tasks, often referred to as ``pretext tasks'', that leverage the inherent structure of video data. Some sample tasks include predicting future frames, determining temporal order, or contrasting different clips from the same video. These tasks encourage models to extract high-level, information-rich features that capture complex temporal dynamics and semantic information directly from raw video data. Such general-purpose representations can be leveraged for a wide range of downstream tasks, including action recognition and anomaly detection, making SSL a key tool for advancing automatic video understanding and analysis.

Predicting future frames based on the past and masked frames based on their context are popular pretext tasks used to train SSL systems for video representation learning \citep{srivastava2015unsupervised, mathieu2015deep, denton2018stochastic, tong2022videomae, girdhar2023omnimae}. These models are generative by nature, as they make predictions in the input pixel space. This approach requires the model to generate all of the low-level details about the target frames, such as textures, object patterns, and background dynamics (e.g., ripples in water or leaves moving in the wind). However, this level of detail can be burdensome and may not be necessary for capturing high-level information, such as the locations and interactions between different objects in a video.

Joint embedding predictive architectures (JEPA) offer a promising alternative to generative models  \citep{lecun2022path, bardes2023v, assran2023self}. Instead of focusing on pixel-level predictions, JEPA models operate at a higher level of abstraction. In particular, in a JEPA, prediction occurs in the abstract representation space. This approach is less computationally expensive compared to pixel-level prediction and can lead to a higher level of abstraction and the ability to eliminate irrelevant details from the target representation \citep{vondrick2016anticipating}. By making predictions in the abstract representation space, the model can ignore unnecessary details and concentrate on the high-level information present in the data. This capability is particularly useful for video data, which is highly redundant, as JEPA can efficiently extract meaningful, high-level representations by focusing on essential temporal and semantic patterns rather than low-level pixel details. Our work focuses on building JEPA models for video representation learning by predicting the hidden representation of a set of future (target) frames from the hidden representation of a set of input frames.

\begin{figure}[t]
\centering
\begin{subfigure}{.48\textwidth}
  \includegraphics[width=\textwidth]{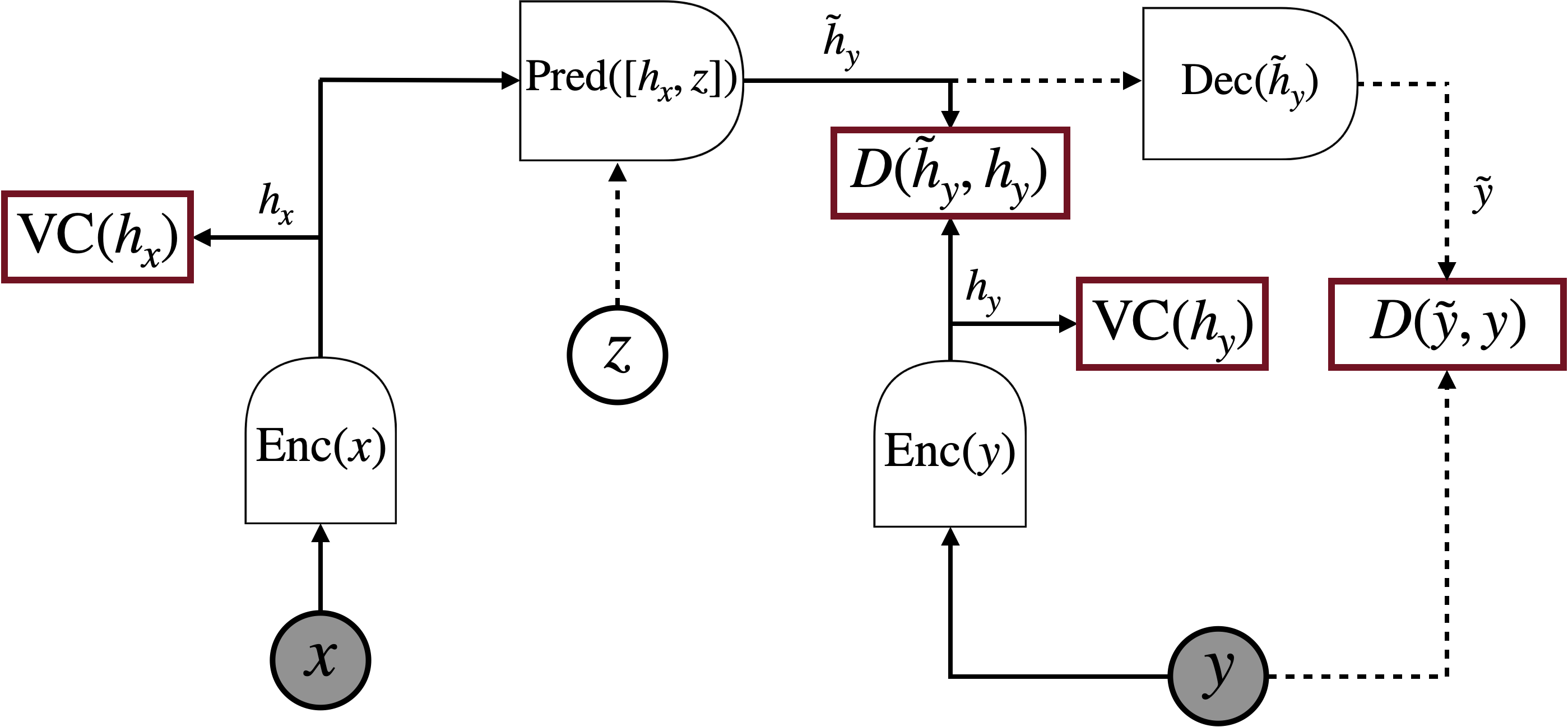}
  \caption{Video JEPA with Variance-Covariance Regularization.}
  \label{fig:JEPA-w-latent}
\end{subfigure}
\hfill
\begin{subfigure}{.48\textwidth}
  \includegraphics[width=\textwidth]{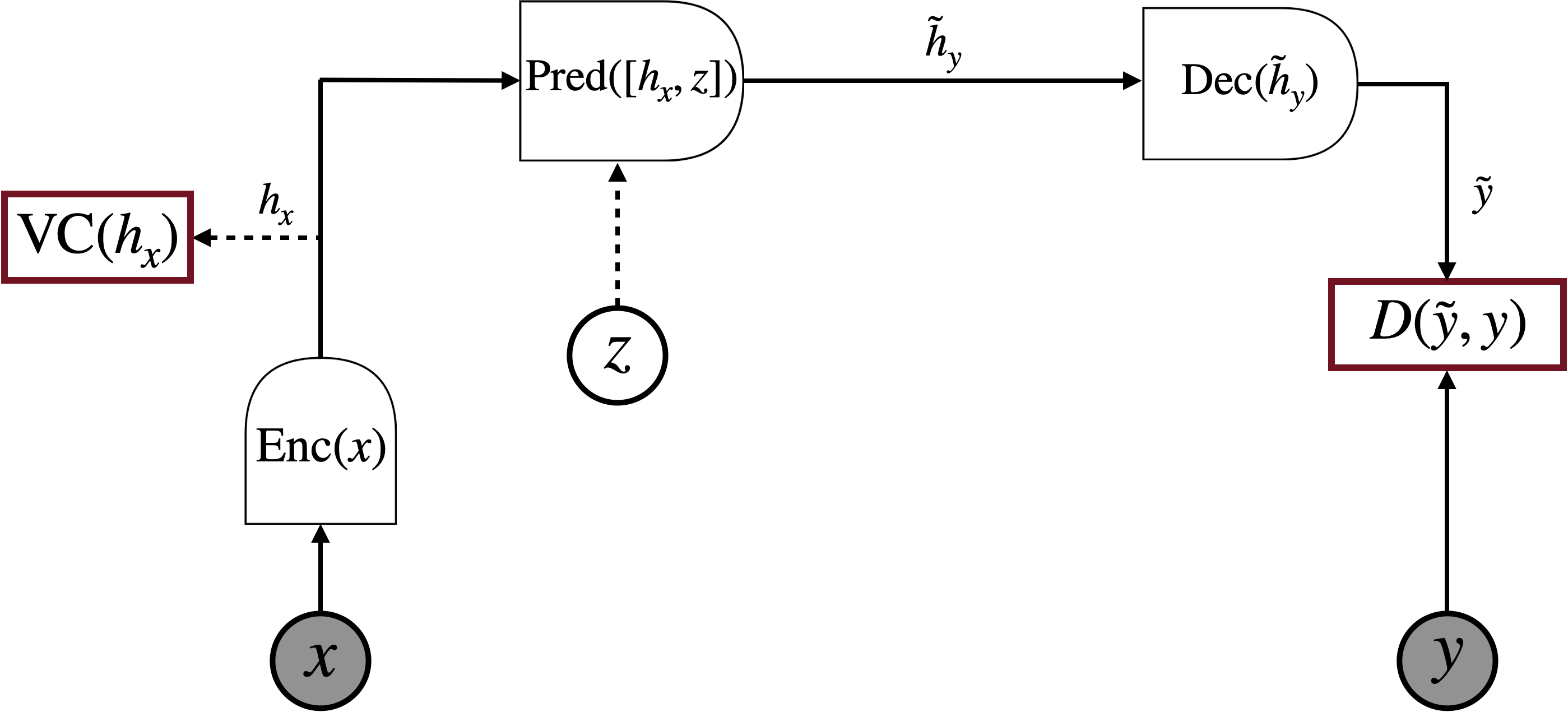}
  \caption{Generative model.}
  \label{fig:GEN-w-latent}
\end{subfigure}
\setlength\belowcaptionskip{-0.5cm}
\caption{Models for self-supervised video representation learning. Inputs $x$ and targets $y$ denote input and target frames coming from the same video, respectively. The optional latent variable $z$ is intended to capture information about the targets $y$ not present in $x$. In the case of VJ-VCR, the Decoder module is optional. $D$ denotes the MSE loss function in the hidden representation space or in the input (pixel) space. $\mathrm{VC}$ denotes variance-covariance regularization. }
\label{fig:models}
\end{figure}

A key challenge in training JEPA models is preventing collapse in the hidden representations. Without taking any precautions, JEPAs are prone to a collapse mode during which the model becomes invariant to the input and maps everything to the same internal representation. Unlike previous approaches that use JEPA for video representation learning, we propose a JEPA model that applies variance-covariance regularization to the model's hidden representations in order to prevent collapse. We call our model \textbf{V}ideo \textbf{J}EPA with \textbf{V}ariance-\textbf{C}ovariance \textbf{R}egularization (\textbf{VJ-VCR}; Figure \ref{fig:JEPA-w-latent}). In essence, variance-covariance regularization encourages the hidden representations of the model to exhibit high variance within each hidden component while simultaneously maintaining low covariance between different hidden components \citep{bardes2022vicreg}. We find that this regularization strategy in the context of video representation learning with JEPA successfully prevents collapse. Moreover, we show empirically that VJ-VCR learns video representations that extract high-level information about the underlying inputs.

One inherent challenge of predicting the future from the past, particularly in real-world settings, is the stochastic nature of the future---it is often not fully predictable based on past information alone. This uncertainty arises from the many possible outcomes that can result from a given context, influenced by factors that may not be directly observable in the past data. To address this challenge in our VJ-VCR setup, we propose introducing latent variables that encode information about the uncertain aspects of the future. These latent variables capture potential variations in the future that cannot be inferred solely from the past, allowing the model to represent and account for the inherent uncertainty in the predictions. By incorporating latent variables, the VJ-VCR  model can better handle stochasticity and generate more robust and realistic representations of the future.

In summary, the main contributions of our work are as follows: 
\begin{itemize}[itemsep=0.05cm,topsep=0pt]
    \item we present VJ-VCR: a JEPA model for video representation learning which utilizes variance-covariance regularization to prevent collapse in the hidden representations,
    \item we demonstrate that representations from VJ-VCR capture high-level information about the input, which is useful for downstream tasks that require an understanding of the underlying dynamics present in the data, 
    \item we show that representations from VJ-VCR outperform those obtained from generative models on several downstream tasks,
    \item we propose ways to incorporate latent variables in the VJ-VCR setup that capture information about uncertainty in the future.
\end{itemize}

Through addressing the issues of collapse and integrating uncertainty into our JEPA model, our approach lays groundwork for future advancements in self-supervised video learning, especially in cases where model efficiency and interpretability are desired.

\section{Related Literature}

\paragraph{Joint-embedding Predictive Architectures} 
\textit{Joint-embedding architectures} (JEA) are a class of self-supervised deep learning models that capture compatibility and dependencies between two inputs \citep{lecun2022path}. The underlying objective of JEA is to assign low energy to compatible inputs and high energy to non-compatible inputs. Examples of such systems are Siamese Networks \citep{becker1992self, NIPS1993_288cc0ff, hadsell2006dimensionality} which contain two identical sub-networks that share a common representation space and are trained to learn a similarity metric between their inputs. Some recent examples of joint-embedding architectures include SimCLR \citep{chen2020simple},  Barlow Twins \citep{zbontar2021barlow}, and VICReg \citep{bardes2022vicreg}. These are invariance-based methods that train an encoder to produce similar representations for different views of the same image. \textit{Joint-embedding predictive architectures} (JEPA), are a type of JEA that incorporates a predictor module in addition to the Siamese encoder \citep{bardes2023v, assran2023self}. More specifically, given two inputs $x$ and $y$, and their corresponding embeddings $h_x$ and $h_y$ from the encoder, the role of the predictor is to learn to predict $h_y$ from $h_x$ for compatible $x$ and $y$. In contrast to generative models that aim to predict the actual target $y$ from input $x$, the JEPA approach to learning dependencies between inputs $x$ and $y$ operates in the abstract representation space. This encourages the model to prioritize learning high-level features over low-level details.

\paragraph{Representation Collapse} One challenge when training joint-embedding architectures is that they are prone to representation collapse: a scenario in which the encoder becomes invariant to its inputs and maps all of them to the same hidden representation \citep{jing2021understanding, shwartz2023information}. This type of collapse is a special case of dimensional collapse, a scenario in which hidden representations lie on a very low-dimensional manifold within the representation space \citep{li2022understanding, jing2021understanding}. There exist different approaches to preventing representation collapse such as: contrastive methods that have a loss that pushes away embeddings belonging to incompatible inputs \citep{chen2020simple}, methods that introduce architectural asymmetries such as momentum encoders \citep{grill2020bootstrap} or non-differentiable operations \citep{chen2021exploring}, information-maximization methods that aim to maximize the entropy of the average representation \citep{caron2021emerging}. Our work is closely related to yet another approach to avoiding representation collapse, namely, de-correlating the representations to eliminate redundancy in the underlying features, as recently promoted in Barlow Twins \citep{zbontar2021barlow} and VICReg \citep{bardes2022vicreg}. In particular, VICReg introduces two regularization terms that, for a given set of sample images and their corresponding embeddings, encourage high variance within each feature component while minimizing covariance between distinct feature components. This design aims to ensure that the learned features are both informative and diverse. In this work, we extend the VICReg regularization paradigm from the image domain to the video domain. We demonstrate that this approach prevents representation collapse in our VJ-VCR model and enables the learning of informative video representations.

\paragraph{Video Representation Learning}
Learning good video representations is an increasingly important topic in computer vision, as it forms the foundation for a wide range of applications, such as action recognition, video captioning, video understanding, and anomaly detection. Popular architectures for video representation learning methods include CNNs \citep{tran2018closer, feichtenhofer2019slowfast} and, more recently, Vision Transformers (ViT) \citep{tong2022videomae, wang2023videomae, arnab2021vivit}. In this work we adopt CNNs as a backbone.
In the self-supervised setting, learning typically happens through pretext tasks such as masked autoencoding, reconstruction, future frame prediction, and frame order detection \citep{denton2018stochastic, piergiovanni2019evolving, tong2022videomae}. In this work, we focus on prediction of the future in the abstract representation space. The recurrent model in \cite{han2019video} learns to predict future frames at the abstract representation level and is trained with a contrastive loss. One limitation of this approach is that it may require a large number of negative samples to extract informative data representations, particularly as the dimensionality of the hidden representations increases. The V-JEPA model \citep{bardes2023v} is most closely related to our VJ-VCR model. V-JEPA makes predictions in the hidden representation space and avoids collapse in the representations by masking the inputs to one of the branches of its Siamese encoder. Additionally, V-JEPA introduces an architectural asymmetry, as one branch of the Siamese encoder is an exponential moving average of the other and is isolated from the rest of the network with a stop-gradient operation. Our work is the first to train a JEPA for video representation learning by utilizing variance-covariance regularization in order to prevent collapse in the hidden representations, without relying on negative samples or architectural asymmetry.

\section{Method}
In the following sections, present our VJ-VCR model: a self-supervised JEPA method for video representation learning that is trained by making predictions in the abstract representation space and that uses variance-covariance regularization to prevent representation collapse.
Prediction in the hidden representation space encourages the model to focus on high-level rather than low-level details, while the variance and covariance regularization is a way to directly ensure that the hidden representations are diverse and informative.

\subsection{The VJ-VCR Model}
Figure \ref{fig:JEPA-w-latent} depicts our VJ-VCR model for self-supervised video representation learning that, given the hidden representation of input frames, is trained to predict the hidden representaiton of future frames. It consists of encoder, predictor, and an optional decoder modules that can have any desired architecture.  VJ-VCR can also incorporate a latent variable to account for stochasticity in the future. 

\textbf{Encoder~} Given a set of input frames $x$ and target frames $y$ coming from the same video, an encoder maps these frames into their corresponding hidden representations $h_x$ and $h_y$, respectively. In order to prevent collapse in the representation space, we apply variance-covariance regularization to the hidden representations $h_x$ and $h_y$, which we cover in detail in \ref{sec:vc-reg}.

\textbf{Predictor~} The predictor takes the hidden state of the input frames $h_x$ 
and predicts the hidden state of the target frames, $\tilde{h}_y$. The predictor can take a latent variable as an input in addition to $h_x$, as described below. 

\textbf{Latent Variable~} VJ-VCR can incorporate a latent variable $z$ to facilitate the prediction task in case that the target frames are not a completely deterministic version of the inputs, i.e.~there are some unobserved variables that influence what the target frames contain but cannot be inferred from the input. Including a latent variable can improve the interpretability of the model by separating deterministic from stochastic information present in the data. Moreover, using a latent variable allows the model to select from a range of potentially many possible future outcomes.

\textbf{Decoder~}  VJ-VCR optionally incorporates a decoder, which is trained to reconstruct the target frames $y$ from the predicted hidden representation of these frames $\tilde{h}_y$.

\subsection{Variance-Covariance Regularization}
\label{sec:vc-reg}

Variance-Covariance Regularization (VCR) is an effective way to  prevent collapse in a JEPA. We adapt its formulation presented in \cite{bardes2022vicreg} and \cite{zhu2023variance} to the setting of video data. Let  $X = \{x_1, x_2, \ldots, x_N\}\subset \mathbb{R}^{T\times H\times W \times N}$ be sets of frames coming from $N$ videos, where $T$ is the number of frames from each video, and $H$ and $W$ denote the height and width of each frame, respectively. Let $f_{\theta}$  be a neural network parameterized by $\theta$ that maps the inputs in $X$ to their (flattened) $d$-dimensional hidden representations $\mathrm{H} = \{h_i~ | ~ h_i = f_{\theta}(x_i)\}_{i=1}^{N}$, where $h_i\in \mathbb{R}^{T\times d}$. VCR's objective is to ensure that the hidden representations in $\mathrm{H}$ exhibit high variance and low covariance. 

In particular, VCR encourages the variance along each of the $d$ components and $T$ time steps to be above a certain threshold $\tau > 0$. This is achieved with a hinge loss regularization term:
\begin{equation}
    \label{eq:VCR-variance-term}
    l_{\mathrm{var}}(\mathrm{H}) = \frac{1}{Td}\sum_{t=1}^{T}\sum_{k=1}^{d}  \max\Big(0, \tau - \sqrt{\mathrm{Var}(\mathrm{H}_{t, k}) + \varepsilon}\Big),
\end{equation}
where $\mathrm{H}_{t,k} = \{h_1^{(t,k)}, h_2^{(t,k)}, \ldots, h_N^{(t,k)}\}\subset \mathbb{R}$ is the set of all $k$-th  components of the representations in $\mathrm{H}$ at time frame $t$, the $\mathrm{Var}(\mathrm{Z})$ function computes the variance of $\mathrm{Z} = \{z_i\}_{i=1}^N\subset \mathbb{R}$ as $\mathrm{Var}(\mathrm{Z}) = \frac{1}{N-1}\sum_{i=1}^{N}(z_i - \overline{z})^2$ using the mean $\overline{z} =\frac{1}{N}\sum_{i=1}^{N}z_i$, and $\varepsilon$ is a small constant introduced for numerical stability. In our experiments we set $\tau = 1$.

The VCR covariance regularization term in our setup is defined by:

\begin{equation}
    \label{eq:VCR-covariance-term}
    l_{\mathrm{cov}}(\mathrm{H}) = \frac{1}{Td}\sum_{t=1}^T\sum_{i\neq j}  \Big[\mathrm{Cov}(\mathrm{H}_{t,:})\Big]_{i,j}^2.
\end{equation}

It sums the squares of the non-diagonal entries of the covariance matrix of $\mathrm{H}_{t,:} \in \mathbb{R}^{d\times N}$, the hidden representations at time step $t$, namely 

\begin{equation}
    \label{eq:VCR-covariance-definition}    
    \mathrm{Cov}(\mathrm{H}_{t,:}) = \frac{1}{N-1}\sum_{i=1}^{N} (h_i^{(t,:)} - \overline{h}^{(t,:)})(h_i^{(t,:)} - \overline{h}^{(t,:)})^\mathrm{T}
\end{equation}
where $\overline{h}^{(t,:)}\in\mathbb{R}^d$ is the mean of all hidden representations, namely $\overline{h}^{(t,:)} = \frac{1}{N} \sum_{i=1}^N h_{i}^{(t,:)}.$ Minimizing the term in $(\ref{eq:VCR-covariance-term})$ encourages the hidden representations to be de-correlated. 

The final formulation of VCR is a weighted sum of the regularization terms in (\ref{eq:VCR-variance-term}) and (\ref{eq:VCR-covariance-term}):

\begin{equation}
    \label{eq:VCR-definition}    
    l_{\mathrm{vcr}}(\mathrm{H}) = \alpha l_{\mathrm{var}}(\mathrm{H}) + \beta  l_{\mathrm{cov}}(\mathrm{H}). 
\end{equation}

In practice, VCR is applied at the batch level during training, rather than the whole dataset, i.e.~$N$ can be interpreted as the batch size.

\subsection{Training and Inference} 

\textbf{Training~} We formulate the training objective of our VJ-VCR model depicted in Figure \ref{fig:JEPA-w-latent} in the framework of energy-based learning \citep{lecun2006tutorial}. An energy function is a function that assigns a scalar value to configurations of observed and latent variables in a given system, with lower levels of energy corresponding to more compatible configurations. In our setting, $x$ and $y$ (the input and target frames, respectively) are the observed variables, and $z$ denotes the unobserved variables. VJ-VCR's objective is to minimize an energy function defined as a weighted sum of several components over the set of training data, namely, the prediction error between the predicted ($\tilde{h}_y$) and the actual ($h_y$) hidden state of the target frames; the variance-covariance regularization terms, which promote diverse and uncorrelated hidden features; and, optionally, the reconstruction error from the decoder:

\begin{align}
\label{eq:vj-vcr-objective-funciton}
    \mathrm{E}_{\theta_{\text{VJ-VCR}}}(x, y, z) & = D(\tilde{h}_y, h_y)  +  l_{\mathrm{vcr}}([h_x, h_y])  + \gamma  D(\tilde{y}, y) \nonumber \\
    & = \|\mathrm{Pred}(h_x,z) - h_y\|_2^2+ \alpha l_{\mathrm{var}}(\mathrm{[h_x, h_y]}) + \beta  l_{\mathrm{cov}}(\mathrm{[h_x, h_y]}) + \gamma  \|\mathrm{Dec}(\mathrm{Pred}(h_x,z))- y\|_2^2 ,
\end{align}

where $\theta_{\text{VJ-VCR}}$ denotes the model parameters, $D$ denotes mean-squared error (MSE), $h_x$ and $h_y$ are the hidden representations of the input and target frames, respectively, and $[h_x, h_y]$ denotes the set of hidden representations for $x$ and $y$. In the rest of the paper, unless otherwise noted, the reconstruction loss is not used during training of VJ-VCR, namely $\gamma = 0$.

\paragraph{Inference} During inference, the optimal value $z^*$ for the latent variable $z$, given set of input frames $x$ and target frames $y$, is obtained by minimizing the energy function defined in \ref{eq:vj-vcr-objective-funciton} with respect to $z$, namely:

\begin{equation}
    \label{eq:infer-z}
    z^{*} =  \argmin_z \mathrm{E}_{\theta_{\text{VJ-VCR}}}(x, y, z) = \argmin_z \big( \|\mathrm{Pred}(h_x,z) - h_y\|_2^2 + \gamma  \|\mathrm{Dec}(\mathrm{Pred}(h_x,z))- y\|_2^2\Big).
\end{equation}

In this work, we adopt gradient-based methods for solving this optimization problem. 

\section{Experimental Setup}
\label{sec:experimental-setup}
In this work, we aim to evaluate our hypothesis that a JEPA-based approach to self-supervised video representation learning, using our proposed VJ-VCR model, can generate video representations that better capture high-level information about the underlying videos than generative-based models, e.g.~by encoding the dynamics of moving objects. For this purpose, we design various  experiments with several datasets and compare VJ-VCR to a generative-based baseline. The details of our experimental setup are outlined in the following subsections.

\subsection{Generative Model Baseline}
Figure \ref{fig:GEN-w-latent} illustrates the generative model for self-supervised video representation learning that we use as a baseline for comparison with our VJ-VCR. The generative model shares the same building blocks as the VJ-VCR model, but it differs in two key aspects: it incorporates a decoder by default, and its training objective is to perform predictions in the input (pixel) space, rather than in the abstract representation space. In particular, the generative model's objective is to minimize the energy function defined as a weighted sum of the reconstruction error from the decoder and, optionally, the variance-covariance regularization term:

\begin{align}
\label{eq:gen-objective-funciton}
    \mathrm{E}_{\theta_{\mathrm{GEN}}}(x, y, z) & =  D(\tilde{y}, y) + l_{\mathrm{vcr}}([h_x, h_y]) =   \|\mathrm{Dec}(\mathrm{Pred}(h_x,z))- y\|_2^2 + \alpha l_{\mathrm{var}}(\mathrm{[h_x, h_y]}) + \beta  l_{\mathrm{cov}}(\mathrm{[h_x, h_y]})
\end{align}

Similarly to VJ-VCR, the generative model can also incorporate latent variables $z$ that encode information about the future frames which is not directly predictable from the past. 

\subsection{Datasets} 
\label{sec:datasets}
We validate our approach to video representation learning with experiments related to understanding the dynamics of moving objects. The datasets that we use for this purpose can be categorized into deterministic and non-deterministic ones depending on whether they inherently contain some stochastic events as described below. Additional dataset details can be found in Appendix \ref{app:datasets}.

\paragraph{Deterministic Setting}
MovingMNIST \citep{srivastava2015unsupervised} is a synthetic dataset that consists of videos of MNIST \citep{lecun1998gradient} digits of size $28\times 28$ moving with randomly chosen constant velocity across a $64\times 64$ black frame. In its original version, when a digit hits a wall, it bounces off the wall in a deterministic fashion. 

The CLEVRER dataset \citep{yi2019clevrer} consists of synthetic videos of colliding objects. Every video frame is annotated with each of the objects’ shape, location, velocity, and collision events. When generating input pairs ($x$, $y$) in our setup, we filter out ones in which new objects appear in the clip $y$, making the data deterministic.

\paragraph{Non-deterministic Setting} 
Our custom stochastic version of MovingMNIST is the following. In the first 3 frames of each video, the digit moves horizontally. In the following 3 frames, the digit randomly switches trajectory in one of five possible directions, namely $\psi \in \{\frac{2k\pi}{5} ~|~ k\in\{0, 1, \ldots, 5\}\}$.

CATER \citep{girdhar2019cater} is a dataset based on CLEVR \citep{johnson2017clevr} designed for spatiotemporal video reasoning tasks. It features moving objects that can interact with each other and perform a set of 14 pre-determined actions. Unlike CLEVRER in which interactions between objects are deterministic, in CATER, objects can randomly begin performing actions within the video. By design, multiple actions can be performed at any given point in the video. Since the actions are chosen at random, the input frames alone cannot deterministically predict what the future frames will contain.

\subsection{Evaluation} 

\label{vj-vcr-eval}
We evaluate the pretrained and frozen VJ-VCR and generative models through several means: 
\begin{itemize}[itemsep=0.05cm,topsep=0cm]
    \item \textbf{Predicting high-level information from hidden representations}: we assess the models' ability to capture high-level dynamic information from the video data, such as object speeds.
    \item \textbf{Predicting high-level information from latent variables}: in non-deterministic settings, we evaluate the models' capacity to utilize latent variables by predicting high-level information, such as actions present in a video, from inferred latent variables.
    \item \textbf{Visualizing learned hidden representations}: we train a decoder to map the hidden representations back to pixel space, allowing us to visualize the information encoded within them.    
    \item \textbf{Information theoretic analysis}: we employ information theoretic metrics to quantify the amount of information contained within the models' hidden representations.
\end{itemize}
Specifically, we outline the evaluation for the deterministic and non-deterministic datasets in the following subsections. Full training details and hyperparameter values $\alpha, \beta, \gamma$ can be found in Appendix \ref{app:vj-vcr-training-details}. Additional evaluation details can be found in Appendix \ref{app:eval-details}.

\subsubsection{Deterministic Setting} 
\label{sec:det-eval}
During evaluation, we would like to check whether information about the underlying high-level dynamics in the video is captured in the predicted hidden representation of the target frames, $\tilde{h}_y$. In particular, we propose to predict the speed $v$ (a scalar value) of an object in the video from $\tilde{h}_y$. For this purpose, we freeze the pre-trained encoder and predictor for models from Figure \ref{fig:models}. We then train a linear regression that takes the predictor's outputs and, in case of MovingMNIST, predict the (constant) speed of the moving digit, or, in the case of CLEVRER, predicts the speed of the fastest object in the last target frame. We refer to this evaluation as \textit{speed probing}. 

\subsubsection{Non-deterministic Setting} 
\label{sec:non-det-eval}

\paragraph{MovingMNIST} In the case of MovingMNIST, we evaluate whether we can isolate stochastic information from deterministic information  in the latent variables $z$ using our VJ-VCR model. Our goal is for the latent variable to encode the stochastic information of the random switch $\psi$ in the trajectory of the digit.  We consider two ways to incorporate the latent variable $z$ into the VJ-VCR setup. 

In the first setting, $z$ is modeled as a discrete latent variable. In particular, it is a one-hot vector of dimension 5 (which is the number of possible random switches in the trajectory by design). The discrete latent $z$ influences the top linear layer of the Predictor, which can be one of 5 options based on the value of $z$. In other words, the active component in $z$ selects the last linear layer of the Predictor. 

In the second setting, we regularize $z$ to be a sparse vector of dimension 20. During inference, given input frames $x$ and target frames $y$, we apply the FISTA algorithm \citep{beck2009fast} to find the sparse $z^*$ that minimizes the energy defined in \ref{eq:vj-vcr-objective-funciton}. We experiment with different values of the sparsity regularization for $z$. One can view the discrete latent variable $z$ as an extreme case of a sparse $z$.

During evaluation of VJ-VCR models that incorporate discrete or sparse latent variables $z$, we train a linear classifier that takes the inferred $z^*$'s as input and predicts their corresponding switches in trajectory $\psi$. Additionally, we are interested in evaluating whether the latent $z^*$'s contain static information such as digit identity in addition to stochastic information about the switch in the digit's trajectory. We use linear probing to predict the digit identity from $z^{*}$. Finally, we train a decoder on top of the predicted frozen hidden representations $\tilde{h}_y$ for the target frames $y$ to visualize the information contained in them. 

\paragraph{CATER} In the case of CATER, we evaluate pre-trained VJ-VCR and generative models using a standard benchmark of multi-label action recognition. To address the stochastic nature of the CATER videos, we incorporate the latent variable $z$ into our self-supervised video representations learning setups as depicted in Figure \ref{fig:models}. We evaluate our models by predicting the aggregate actions $a_y$ present in the future target frames $y$ from the inferred latent variable $z^*$ associated with the target frames. In particular, during pre-training, the latent $z$ is a binary vector that provides the ground truth actions across all time steps to the predictor. During inference, for each sample, we use several iterations of gradient descent to compute $z^{*}$, an approach similar to the algorithm described in \cite{henaff2017prediction}. We then evaluate whether the inferred latent variable $z^{*}$ captures the aggregated set of ground-truth actions $a_y$ present in the target frames through linear probing for the task of multi-label action recognition. As evaluation metric, we report the mean average precision (mAP) measured on the validation set following \citep{girdhar2019cater}. The average precision per-class $c$ is defined as the ratio between true positive and the sum of true positive and false negative predictions, $\mathrm{AP}_{c} = \frac{\mathrm{TP}_{c}}{\mathrm{TP}_{c} + \mathrm{FP}_{c}}$ and mAP is computed by taking the average over all classes: $\mathrm{mAP} = \frac{1}{|C|}\sum_{c\in C}\mathrm{AP}_{c}$. 


\subsection{Architecture}
In experiments with MovingMNIST, we model the encoder as a 5-layer convolutional neural network with batch norm and ReLU activation function at each layer. It has 3 spatial and 2 temporal convolutions followed by an average pooling layer. The predictor is an MLP with 2 hidden layers. Unless otherwise noted, the input $x$ contains 3 frames and the target $y$ contains the following 12 frames in the video. The predictor outputs the hidden state of the 12 target frames $y$ simultaneously. 

In experiments with CLEVRER and CATER, we use a SimVP encoder \citep{gao2022simvp} and a Swin Transformer \citep{liu2021swin} for the predictor. Unless otherwise noted, the input $x$ contains 6 frames and the target $y$ contains the following 20 frames in the video for CLEVERER. For CATER, we subsample the video at the rate 8 frames per second following \citep{girdhar2019cater} and feed 50 frames as input and the following 50 frames as target. 

In all experiments that use a decoder, its architecture mirrors that of the encoder. It takes the predicted hidden state of the target frames $\tilde{h}_y$ as input to reconstruct the target frames $y$. The decoder can be trained simultaneously with the rest of the system or, alternatively, it can be trained on top of the pretrained and frozen encoder and predictor.

\subsection{Implementation and Hardware}
For our experiments, we use the publicly available PyTorch \citep{paszke2019pytorch} codebase OpenSTL \citep{tan2023openstl} which can be found here: 
\url{https://github.com/chengtan9907/OpenSTL}. 
We train our models on one NVIDIA RTX 8000 GPU card and all of our experiments take less than 48 hours to run.

\begin{table}[t]
\centering
\caption{Evaluation of self-supervised models trained on MovingMNIST and CLEVRER with two different types of losses, namely prediction loss in the hidden space, Loss$(h_y)$, and reconstruction loss in pixel space, Loss$(y)$, and with or without variance-covariance regularization (VCR). In the case of MovingMNIST, we report the MSE of a linear regression predicting the speed of the moving digit and the average reconstruction quality in terms of PSNR. In the case of CLEVRER, we report the MSE of a linear regression predicting the speed of the fastest object in the last predicted frame as well as the estimated rank of the learned hidden representations in terms of RankMe. All metrics are measured on the validation set.}
\label{tab:mmnist-clevrer-speed}
\vskip\baselineskip 
    \begin{tabular}{cccccccc}
    \toprule
  \multirow{2}{*}{Model} &  \multirow{2}{*}{Loss($h_y$)} & \multirow{2}{*}{Loss($y$) } & \multirow{2}{*}{VCR} & \multicolumn{2}{c}{MovingMNIST} & \multicolumn{2}{c}{CLEVRER} \\
 & &  &  & MSE  in $v$ $\downarrow$ & PSNR $\uparrow$ & MSE  in $v$ $\downarrow$ & RankMe $\uparrow$ \\
     \toprule
   VJ-VCR w/o Decoder & \cmark & \xmark & \cmark & 0.04 & 19.5 & 0.19  & 423.7 \\
   VJ-VCR with Decoder & \cmark & \cmark & \cmark & 0.04 & 21.2 & 0.19  & 359.8 \\
   Generative with VCR & \xmark & \cmark & \cmark & 0.10 & 22.9 & 0.22  & 427.4 \\
   Generative w/o VCR & \xmark & \cmark & \xmark & 0.15 & 22.8 & 0.23  & 160.2 \\
    \bottomrule
    \end{tabular}
\end{table}

\section{Results}
\label{sec:results}

In this section, we report our findings on the ability of VJ-VCR and generative video representation learning methods to capture information about the dynamics of moving objects using the experimental setup outlined in section \ref{sec:experimental-setup}. Our hypothesis is that VJ-VCR models, which make predictions in the abstract representation space, are better suited for capturing object dynamics (such as their speeds and actions they perform) than generative models. This is because generative models prioritize the reconstruction of low-level pixel details in the target frames, potentially limiting their ability to capture high-level dynamic information.

\subsection{Deterministic Setting: Speed Probing}

In this set of experiments, we use the deterministic version of MovingMNIST and the CLEVRER datasets introduced in section \ref{sec:datasets} and evaluate VJ-VCR and generative models on the task of speed probing as described in section \ref{sec:det-eval}. 

\paragraph{MovingMNIST (deterministic)} Table \ref{tab:mmnist-clevrer-speed} presents the evaluation results on speed probing of four self-supervised models for video representation learning pre-trained with different energy loss functions. As shown in equation \ref{eq:vj-vcr-objective-funciton}, the energy function can incorporate JEPA-style prediction error in the hidden space paired with VCR, and, optionally, reconstruction error in the pixel space. In terms of evaluation, we report the average reconstruction quality achieved by a decoder (measured by PSNR), and the MSE measured during speed probing, i.e.~the MSE of a linear regression trained to predict the speed of the MNIST digit in each video from $\tilde{h}_y$. Both metrics are measured on the validation set. Appendix \ref{app:mnist-visulaization} provides a visualization of the reconstructions obtained from the model in the second row, VJ-VCR simultaneously trained with a decoder, and the model in the forth row, generative one trained without VCR. 
 
The VJ-VCR models in the top two rows achieve the lowest speed probing MSE of 0.04. This indicates that JEPA models trained to make predictions in the abstract representation space outperform purely generative models at capturing the dynamics of moving digits. Since the VJ-VCR model in the top row does not make predictions in the pixel space, we separately train a decoder to reconstruct the target images $y$ from the frozen hidden representations $\tilde{h}_y$ of this model. This decoder has the lowest reconstruction quality with PSNR of 19.5, suggesting that some reconstruction details are absent in the hidden representations. The VJ-VCR model in the second row, trained with prediction error in the pixel space, achieves better reconstruction quality of 21.2. This demonstrates that incorporating a reconstruction loss term during JEPA training can enhance the reconstruction quality without compromising the model's ability to predict the underlying video dynamics. In contrast, the model in the third row trained solely with pixel loss and VCR achieves the highest reconstruction quality of 22.9 but performs worse at predicting speed compared to the JEPA-based models with an MSE of 0.10. Finally, the purely generative model in the last row, trained without VCR, closely matches the reconstruction quality of the third model at 22.8 but has the worst speed probing performance with MSE of 0.15. 

These results suggest that prediction in the hidden space can enhance the representations with information about the underlying dynamics of the moving digits. Furthermore, applying VCR to the hidden representations can result in encoding more information about the high level details in the underlying data. Additionally, depending on the downstream application, including a reconstruction loss term in addition to a prediction term in the hidden space can help with retaining the finer details needed to decode the hidden representations into pixel space without sacrificing the higher-level information about dynamics.

\begin{figure}[t]
\centering
\includegraphics[width=0.45\textwidth]{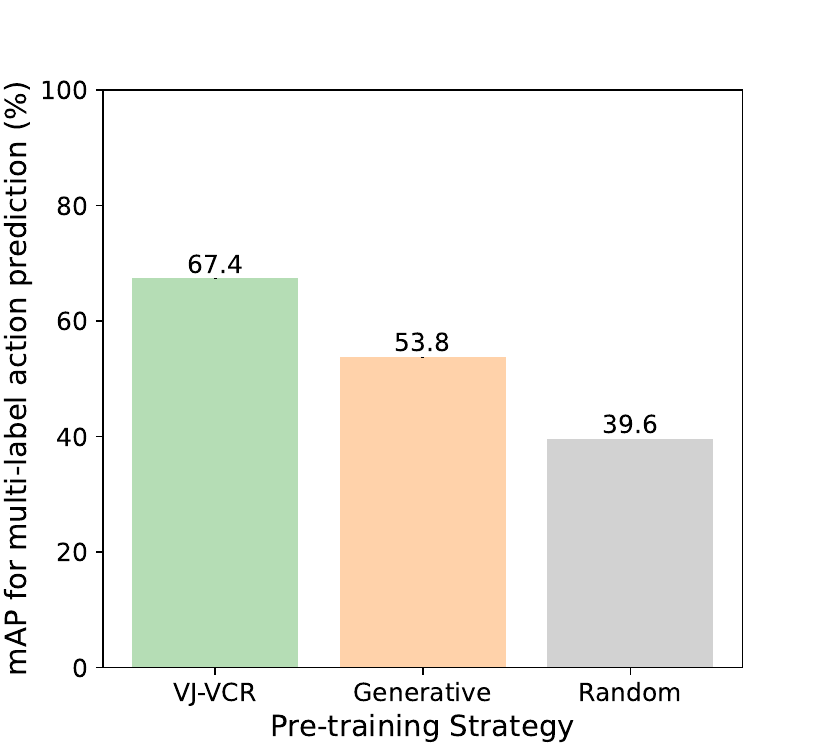}
\caption{Multi-label action recognition performed on the CATER dataset. The aggregated set of actions $a_y$ in the target frames is predicted from the inferred latent variable $z^{*}$ using a linear classifier. Latent variables $z^{*}$ computed from our VJ-VCR pre-trained model are more informative about the underlying actions than those from the pre-trained generative-based models using mAP as an evaluation metric on the validation set. The performance of a linear classifier trained on top of randomly generated latent variables $z^{*}$ in this multi-label setting is $39.6\%$.}
\label{fig:CATER-mAP}
\end{figure}

\paragraph{CLEVRER}

Table \ref{tab:mmnist-clevrer-speed} demonstrates that JEPA-based models (in the top two rows), which utilize prediction loss in the hidden representation space, outperform generative models (in the bottom two rows) in probing the speed of the fastest object in the final target frame of a CLEVRER video. This supports our hypothesis that prediction in the abstract representation space can lead to hidden representations that contain more high-level information about the inputs than those from generative models. 

Additionally, we evaluate the informational content of representations from models pre-trained on CLEVRER using RankMe \citep{garrido2023rankme}, a metric that estimates the effective rank of embeddings produced by joint embedding self-supervised learning methods. Among the JEPA-based models in the top two rows, the one trained with a decoder achieves a lower RankMe score (359.8) compared to the one without (423.7). For generative models, the inclusion of VCR regularization results in a higher RankMe score (427.4) than training without regularization (160.2). Notably, the generative model trained without regularization has the lowest RankMe score among all models. While a higher RankMe score does not necessarily correlate with better speed probing performance, it is noteworthy that adding a reconstruction error term to VJ-VCR's objective reduces the estimated rank, whereas incorporating VCR regularization in the generative model increases it.

\subsection{Non-deterministic Setting: Action Recognition}
In this set of experiments, we use the stochastic version of MovingMNIST and the CATER datasets introduced in section \ref{sec:datasets} and perform evaluation through action recognition using the inferred latent variables $z^*$ as described in section \ref{sec:non-det-eval}. 

\paragraph{MovingMNIST (non-deterministic)}
Table \ref{tab:mmnist-z-switch} summarizes the results on how well the inferred latent $z^*$ can predict the switch in a digit's trajectory $\psi$. We observe that the discrete latent variables can predict the trajectory switch with 80\% accuracy on average. Moreover, discrete latents predict the digit identity at random, as expected, since the switch in trajectory is random by design. 

For sparse latent variables, we observe that the level of sparsity regularization influences both the amount and type of information encoded in them. Namely, we note that at a high level of sparsity (where 80\% of components in the latent variables are 0s, on average) the accuracy of predicting the switch is 94.7\%, In contrast, at lower level of sparsity (20\%), the accuracy increases to 99.5\%. This suggests that higher levels of sparsity reduce the ability to predict the stochastic information about the target frames. 

At the same time, the sparse latents contain non-random amount of static information about the target frames such as digit identity. The ones with higher level of sparsity can predict the digit correctly 31.6\% of the time while for the less sparse ones this accuracy increases to 57.6\%. This suggests that information about the digit identity can ``leak'' into the sparse latent variables. The amount of ``leaking'' can be controlled with the strength of sparsity regularization. However, this comes with a trade-off:higher sparsity levels lead to reduced accuracy in predicting the trajectory switch. Exploring methods to effectively constrain the latent variables to encode only stochastic information, while excluding static information, remains an interesting direction for future research.

\begin{table}[t]
\centering
\caption{Evaluation of latent variables from VJ-VCR pre-trained on our non-deterministic MovingMNIST dataset. We report the average accuracy of predicting switches in trajectory $\psi$ from the inferred latent $z^*$ on the validation set in the case when $z$ is modeled as a discrete or a sparse latent variable with different levels of sparsity regularization. We also report the accuracy of a linear classifier trained to predict the digit identity from the inferred $z^*$.}
\label{tab:mmnist-z-switch}
\vskip\baselineskip 
    \begin{tabular}{ccc}
    \toprule
    Latent Type & $z^* \to \psi$ & $z^* \to~$digit  \\
     \toprule
    discrete (one-hot vector of dimension 5) & 79.7\% & 11.4\%\\
    sparse: high (80\%) sparsity & 94.7\% & 31.6\%\\
    sparse: low (20\%) sparsity & 99.5\% & 57.6\%\\
    \bottomrule
    \end{tabular}
\end{table}

\begin{figure}[t]
    \centering
    \begin{subfigure}{0.8\textwidth}
    \includegraphics[width=\textwidth]{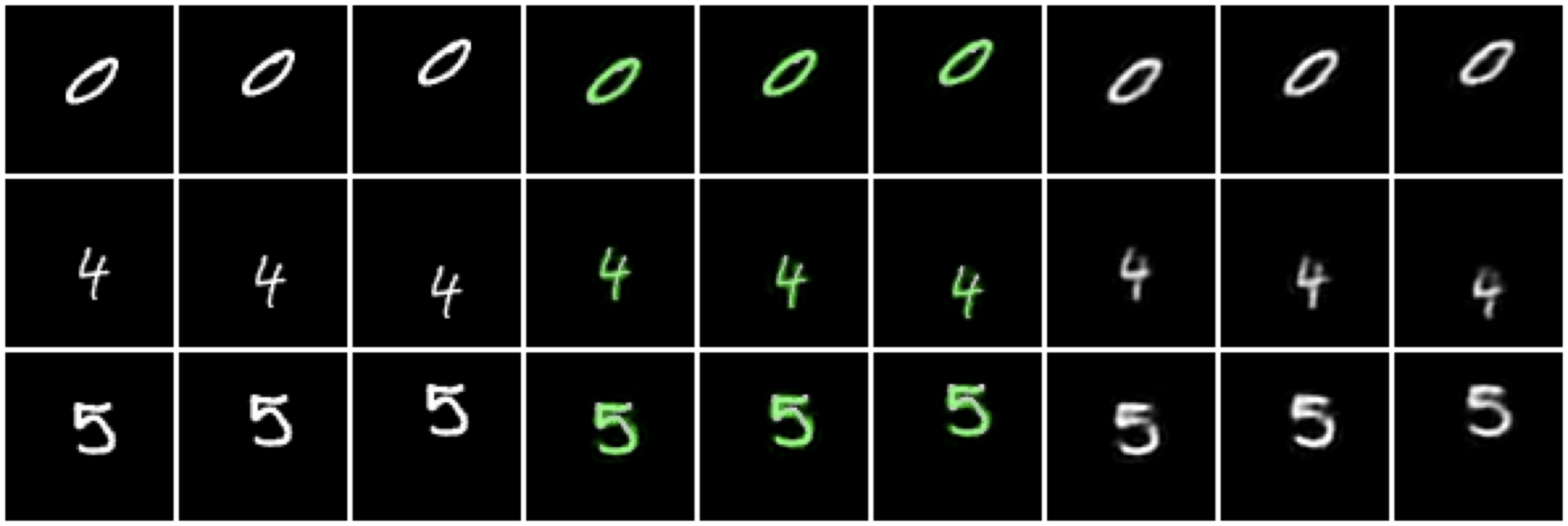}
    \caption{Reconstructions from a VJ-VCR model trained with a latent variable.}
    \label{fig:mmnist-jepa-latent}
    \end{subfigure}
    \hfill
    \begin{subfigure}{0.8\textwidth}
        \includegraphics[width=\textwidth]{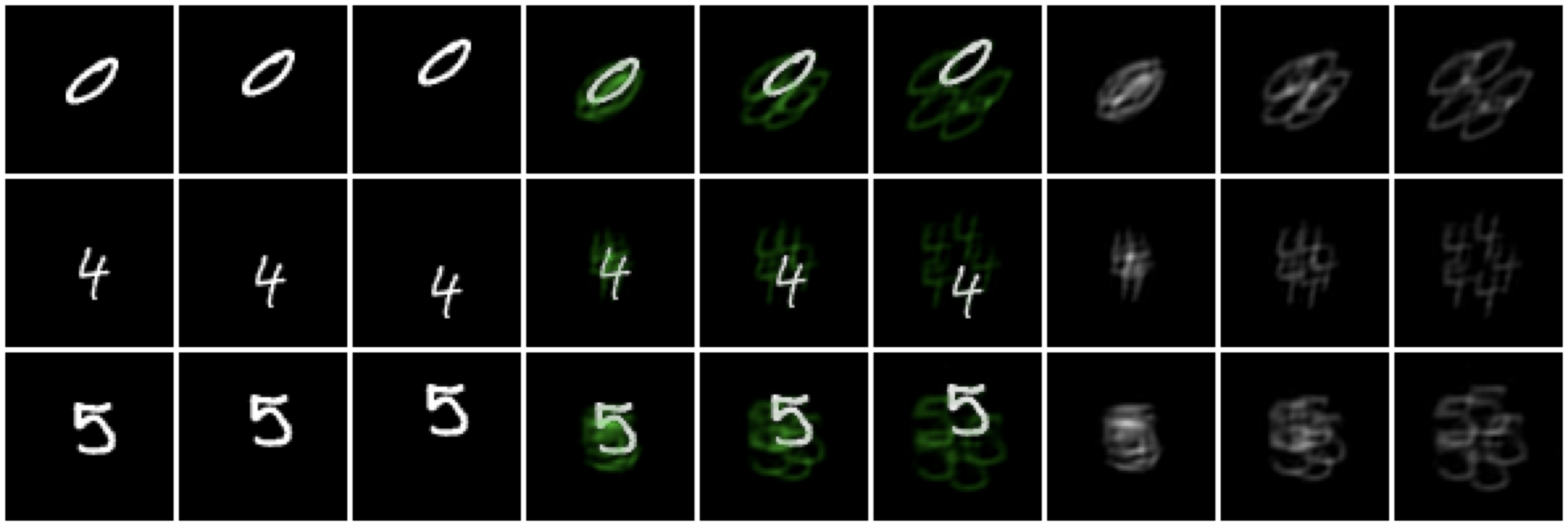}
        \caption{Reconstructions from a VJ-VCR model trained without a latent variable.}
        \label{fig:mmnist-jepa-no-latent}
    \end{subfigure}
    \caption{Reconstructions from our VJ-VCR model trained with and without a latent variable on the MovingMNIST dataset with a random switch in the digit trajectory after the third frame. The first three columns show the original target frames, the last three columns show the model's predictions for the target frames and the middle three columns show the overlap between the original and predicted frames (the latter are displayed in green). The model that does not incorporate a latent variable predicts all possible switches in trajectory of the digit, while the one that uses a latent variable can correctly identify the actual switch in digit trajectory.}
    \label{fig:mmnist-jepa-decoder}
\end{figure}

Additionally, we visualize the information encoded in the predicted hidden representation $\tilde{h}_y$ of the target frames for our VJ-VCR model trained with and without latent $z$ in Figure \ref{fig:mmnist-jepa-decoder}. We note that the model predictions accurately depict the ground truth when a latent variable is used. However, in the absence of a latent, the model cannot choose a particular switch in trajectory and predicts all the possible switches simultaneously. 
These experiments suggest that latent variables can effectively be incorporated in the JEPA framework to encode uncertainty in the future.

\paragraph{CATER} 

In this set of experiments, we evaluate how effectively latent variables from pre-trained VJ-VCR and generative models capture information about future events using the CATER dataset of moving objects. For this purpose, we consider the task of multi-label action recognition from inferred latent variables $z^*$ as described in section \ref{sec:non-det-eval}. As displayed in Figure \ref{fig:CATER-mAP}, our experiments suggest that VJ-VCR pre-trained models outperform generative-based models on this evaluation  by $13.6\%$: the former has mAP of 67.4\% while the latter has mAP of 54.8\%. Both of these results have standard deviation smaller than $2\times 10^{-2}$ for 3 random seeds.  As a reference, the performance of a linear classifier trained on top of randomly generated $z^{*}$ in this multi-label setting is $39.6\%$. 

This supports our hypothesis that making predictions in the hidden representation space, rather than the input space, encourages VJ-VCR models to focus on high-level information about video events, leading to improved performance. While these results are promising, exploring alternative methods of incorporating latent variables within the current framework may yield even better results. We leave this line of research for future work. 

\begin{figure}[t]
    \centering
    \begin{subfigure}{0.48\textwidth}
    \includegraphics[width=\textwidth]{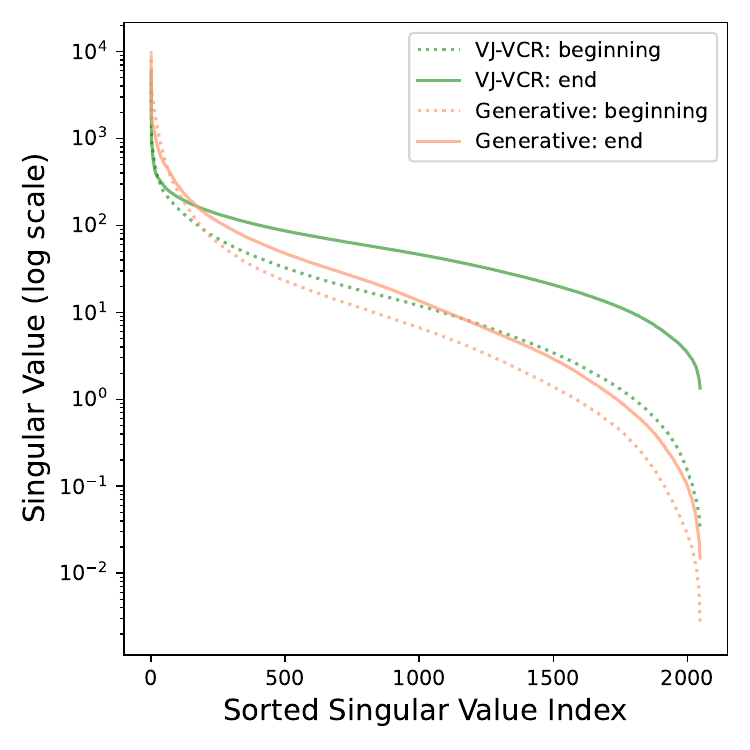}
    \caption{Distribution of the singular values of the hidden representations over the validation set coming from a VJ-VCR and generative-based models pre-trained on CATER in a self-supervised way. The singular values of the VJ-VCR model are more uniformly distributed than those of coming from the generative-based model.}
    \label{fig:svd-over-time}
    \end{subfigure}
    \hfill
    \begin{subfigure}{0.48\textwidth}
        \includegraphics[width=\textwidth]{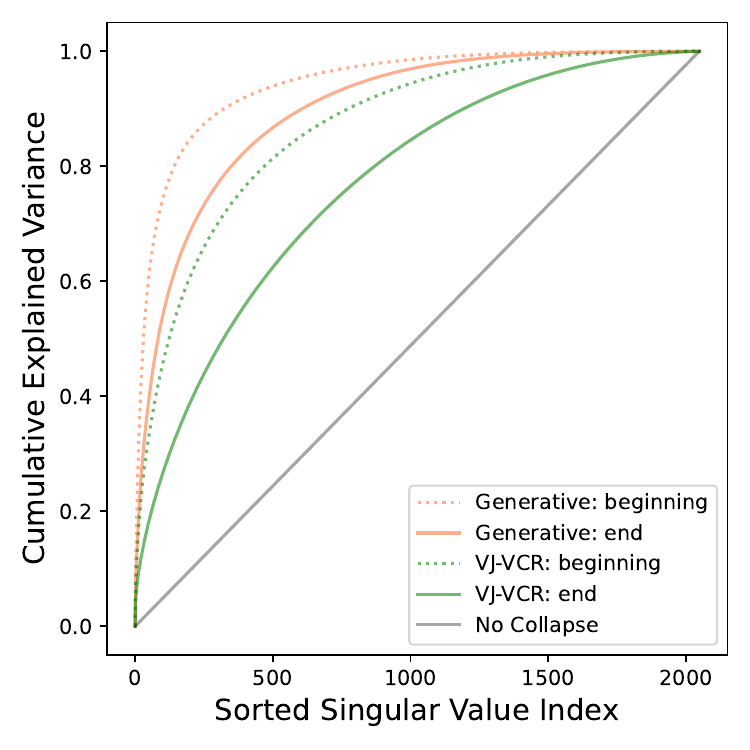}
        \caption{Cumulative explained variance of hidden representations coming from a VJ-VCR and a generative model at different points during training (beginning, middle, and end). The curve of the VJ-VCR model at the last epoch is rising slower than that of the generative-based model, indicating a lower level of dimensional collapse.}
        \label{fig:svd-cumulative-over-time}
    \end{subfigure}
    \caption{Analysis of the informational content of the learned hidden representations of a VJ-VCR and a generative model through singular value decomposition.}
    \label{fig:svd-analysis}
\end{figure}

\section{Analyzing the Information Content of the Learned Representations}

In this section, we analyze the extent of representation collapse in the VJ-VCR and generative models pre-trained through self-supervision on the CATER dataset. In particular, following the approach in \cite{li2022understanding}, for each model we consider the singular value decomposition (SVD) of the matrix $\mathrm{H} \in \mathbb{R}^{N\times d}$ of the encoder's outputs on the validation dataset, where $N$ is the number of validation samples and $d$ is the dimension of the encoder's hidden representations. We track the evolution of the singular values' distribution throughout training. 
The intuition behind it is that having a few dominant singular values would imply that the hidden representations occupy a low-dimensional manifold within $\mathbb{R}^d$. If the distribution of the singular values is more balanced, then the hidden representations have a higher intrinsic dimensionality. 

Figure \ref{fig:svd-over-time} shows the distribution of the singular values of the hidden representation matrix $\mathrm{H}$ for VJ-VCR and a generative model pre-trained on CATER, measured at the beginning and the end of training. The singular values are sorted in descending order.For both models, the distribution becomes more balanced by the end of training. However, the VJ-VCR model exhibits a more balanced singular value distribution compared to the generative model. This observation is further supported by Figure \ref{fig:svd-cumulative-over-time} in which the cumulative explained variance of the VJ-VCR model is increasing more gradually than that of the generative-based model. These results suggest that VJ-VCR pre-training avoids dimensional collapse more effectively than training with a generative objective.

\section{Conclusion}
\label{sec:conclusion}
In this paper, we demonstrate that JEPA-style video representation learning can produce more informative video representations when compared to generative models in the self-supervised setting. Specifically, we apply variance-covariance regularization solely to the top layer of the encoder to prevent representation collapse. Future work could explore extending this regularization to multiple layers of the neural network architecture, potentially enhancing the quality of the learned hidden representations. One limitation of our study is its focus on relatively small synthetic datasets. However, we believe that the proposed VJ-VCR model for video representation learning can generalize effectively to larger and more realistic datasets. Additionally, since JEPA models are trained in the hidden representation space, this can significantly reduce computational costs when compared to generative models, especially in high-dimensional settings. This work contributes to the broader pursuit of developing efficient and reliable AI systems, with the potential to advance a range of applications that require an understanding of complex video data.


\bibliography{main}
\bibliographystyle{tmlr}

\newpage
\appendix
\onecolumn

\section{Datasets}
\label{app:datasets}

\paragraph{MovingMNIST} For experiments with MovingMNIST, we split the original MNIST dataset into 55,000 training and 5,000 validation samples. In the deterministic version of the MovingMNIST, given a sample from the MNIST dataset, we generate a 20-frame  video by randomly sampling the digit's initial location on a $64\times 64$ black canvas and the digit's velocity (which remains constant throughout the video). In the stochastic version of MovingMNIST, given a sample from the MNIST dataset, we generate a 6-frame video as follows. In the first 3 frames of the video, the digit starts at the center of the $64\times 64$ canvas and moves horizontally to the right. In the following 3 frames, the digit randomly switches trajectory in one of five possible directions, namely $\psi \in \{\frac{2k\pi}{5} ~|~ k\in\{0, 1, \ldots, 5\}\}$.

\paragraph{CLEVRER}
The CLEVRER (CoLlision Events for Video REpresentation and Reasoning) dataset consists of synthetic videos of colliding objects \citep{yi2019clevrer}. Each video is 5 seconds long and contains 128 frames with resolution 480 x 320. In our experiments with CLEVRER, we re-shape the videos to $64\times 64$ resolution. Furthermore, we use the official training and validation splits provided at \url{http://clevrer.csail.mit.edu/}.

\paragraph{CATER} CATER \citep{girdhar2019cater} is a synthetic dataset of moving objects that can move independently and also interact with each other. Each video contains 300 frames at 24 frames per second at 320x240 resolution. There are 14 possible actions objects can perform and multiple actions can be present in a single video. In our experiments, we reshape the video to $128\times 128$ resolution. We use a fixed sampling rate of 8 frames per second following the the atomic action recognition setting in  \cite{girdhar2019cater}. Furthermore, we use the pre-generated max2action version of the dataset with only 2 objects moving in each time segment as described in \url{https://github.com/rohitgirdhar/CATER/tree/master/generate}. We use the provided training and validation splits for the max2action version of the dataset.

\section{Training and Evalutation Details}

In all our experiments, all hyperparameters values are chosen through grid search based on the best loss performance on the validation set. 

\subsection{Training Details}
\label{app:vj-vcr-training-details}
For MovingMNIST experiments, we train models for 100 epochs and pick the best one in terms of the self-supervised loss performance on the validation set. For both VJ-VCR and generative models we use a learning rate of $1\mathrm{e}{-3}$ and the Adam optimizer \citep{kingma2014adam} and a batch size of 256. In the deterministic setting, the models are trained to take 3 frames as input and predict  the following 12 frames as output. In the non-deterministic setting, the models are trained to take 3 frames as input and predict the following 3 frames as output. In the case of generative models, we use weight decay of $1\mathrm{e}{-6}$. In VJ-VCR experiments, the variance and covariance regularization coefficients $\alpha$ and $\beta$ are set to 0.5 and 0.1, respectively. In generative-baseline experiments, the variance and covariance regularization coefficients can optionally be set to 0.5 and 0.1, respectively.

For CATER and CLEVRER, we train models for a maximum of 20 epochs and pick the best one in terms of the self-supervised loss performance on the validation set. For both VJ-VCR and generative models we use a learning rate of $1\mathrm{e}{-3}$ and the Adam optimizer. We use a batch size of 256 and 160 for CLEVRER and CATER, respectively.

In the case of CLEVRER, the models are trained to take 6 randomly selected consecutive frames from a video in the training set as input and predict the following 20 frames as output. In VJ-VCR experiments with CLEVRER, the variance and covariance regularization coefficients  $\alpha$ and $\beta$ are set to 1 and 0.1, respectively. In generative-baseline experiments, the variance and covariance regularization coefficients can optionally be set to 1 and 0.1, respectively.

In the case of CATER, the models are trained to take 50 frames from a video as input and predict the following 50 frames as output while subsampling the original video from 24 frames per second to 8 frames per second following \cite{girdhar2019cater}. In VJ-VCR experiments with CATER, the variance and covariance regularization coefficients are set to 1 and 0.4, respectively. In generative-baseline experiments, the variance and covariance regularization coefficients can optionally be set to 1 and 0.4, respectively.

\subsection{Evaluation Details}
\label{app:eval-details}
During evaluation though \textit{speed probing} (see section \ref{sec:det-eval}) with MovingMNIST and CLEVRER, we train a linear regression that takes the predicted hidden representations of the target frames from pre-trained VJ-VCR or generative models and outputs a scalar number for the speed of the desired object (the moving digit in the case of MovingMNIST or the fastest moving object in the last predicted frame in the case of CLEVRER). We use MSE loss and the Adam optimizer with a learning rate of $1\mathrm{e}{-3}$ and for batch size of 256 in all experiments except for the ones with CLEVRER and generative-based models in which case we use a learning rate of $1\mathrm{e}{-4}$.

For experiments with the stochastic version of MovingMNIST and a sparse latent variable $z$ (see section \ref{sec:non-det-eval}) of dimension 20, we train a linear classifier that predicts the (discrete) switch in trajectory $\psi$ from the inferred $z^*$ for each video. We use a batch size of 256 and the Adam optimizer with a learning rate of $1\mathrm{e}{-3}$.

During multi-label classification for action recognition with the CATER dataset (see section \ref{sec:non-det-eval}), we train a linear classifier that takes as input the inferred latent variable $z^{*}\in \mathbb{R}^{|A|\times |y|}$ encoding the actions in the target frames $y$ for each video and outputs the probabilities for each action in the set of possible actions $A$ being present in the target frames. We use BCEWithLogitsLoss from pytorch as our loss functional, the Adam optimizer with a learning rate of $5\mathrm{e}{-4}$, and a batch size of 160.
 
\section{Additional Visualizations on MovingMNIST}
\label{app:mnist-visulaization}
\begin{figure*}[t]
     \centering
     \includegraphics[width=\textwidth]{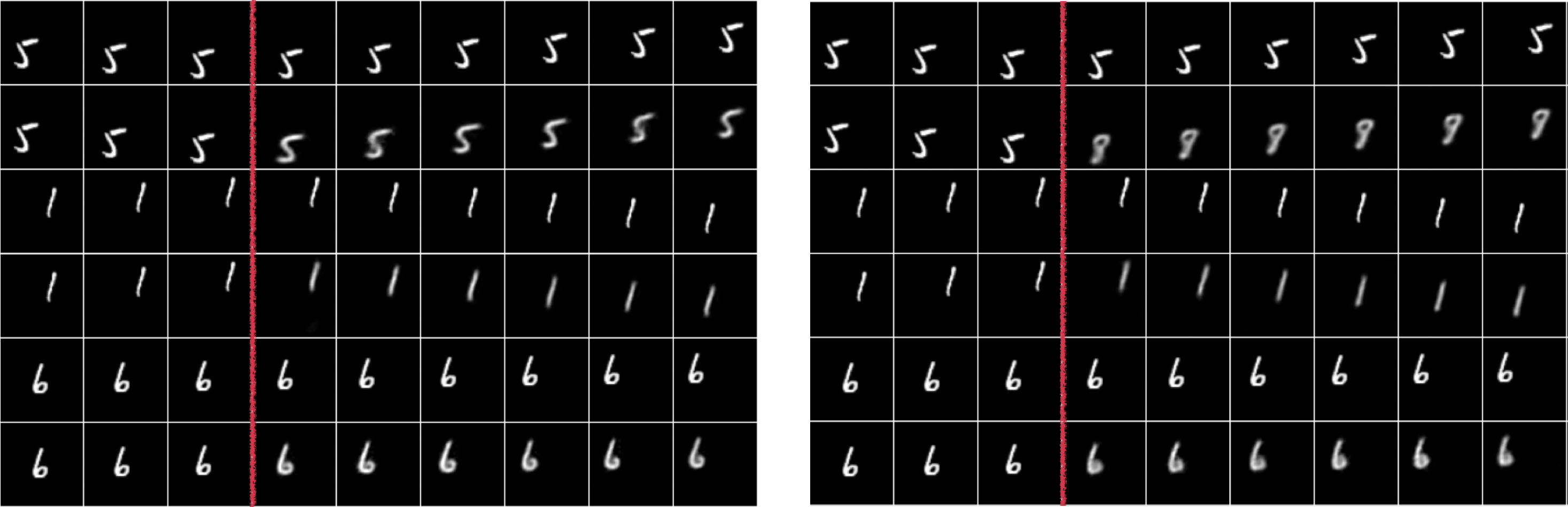}
     \caption{Reconstructions from a generative model trained only with loss in pixel space  (left) and a VJ-VCR model trained with loss in pixel space, loss in the hidden representation space, and variance-covariance regularization (right). Odd rows display 9 ground truth frames. Even rows display the first 3 ground truth frames which are the input to the model followed by the first 6 (out of 12)  reconstructed frames. The model on the left has PSNR of 22.8 and the one on the right has PSNR of 21.2. Both models can predict the trajectories of the digits. Hidden representations from the VJ-VCR model can be used to predict the actual speed of the digits more accurately.}
     \label{fig:MNIST-comparison}
\end{figure*}

As a visual reference, Figure \ref{fig:MNIST-comparison} shows reconstructions from the model trained with reconstruction loss only and from the VJ-VCR model trained with prediction loss in the hidden representation space, reconstruction loss in the pixel space, and variance-covariance regularization. Even though the first model has better reconstruction quality, the second one produces hidden representations which can predict the speed of the moving digits more accurately.

\end{document}